\documentclass[10pt,twocolumn,letterpaper]{article}

\usepackage{iccv}
\usepackage{times}
\usepackage{epsfig}
\usepackage{graphicx}
\usepackage{amsmath}
\usepackage{amssymb}
\usepackage{latexsym}

\usepackage{xspace}
\usepackage{booktabs}       
\usepackage{enumitem}
\usepackage{bm}

\usepackage{url}

\usepackage[breaklinks=true,bookmarks=false]{hyperref}

\iccvfinalcopy 

\ificcvfinal\fi

\usepackage{amsmath,amsfonts}
\usepackage{amssymb,amsopn}
\usepackage{bm} 

\usepackage{amssymb}
\usepackage{amsmath,amsfonts}
\usepackage{amsthm,amsopn}

\usepackage{bm} 

\newcommand{\vct}[1]{\boldsymbol{#1}} 

\newcommand{\ProbOpr}[1]{\mathbb{#1}}

\newcommand{\expect}[2]{%
\ifthenelse{\equal{#2}{}}{\ProbOpr{E}_{#1}}
{\ifthenelse{\equal{#1}{}}{\ProbOpr{E}\left[#2\right]}{\ProbOpr{E}_{#1}\left[#2\right]}}} 
\newcommand{\var}[2]{%
\ifthenelse{\equal{#2}{}}{\ProbOpr{VAR}_{#1}}
{\ifthenelse{\equal{#1}{}}{\ProbOpr{VAR}\left[#2\right]}{\ProbOpr{VAR}_{#1}\left[#2\right]}}} 

\newcommand{\vv}{\vct{v}}

\newcommand{\eat}[1]{}

\begin{document}

\title{Visual Storytelling via Predicting Anchor Word Embeddings in the Stories}

\author{Bowen Zhang$^{1}$ \quad \quad Hexiang Hu$^{1}$ \quad \quad Fei Sha$^{2}$   \\
$^{1}$University of Southern California, $^{2}$ Google AI \\
\texttt{zhan734@usc.edu, hexiangh@usc.edu, fsha@google.com\thanks{\it{On leave from U. of Southern California (feisha@usc.edu)}}}}

\maketitle
\ificcvfinal\thispagestyle{empty}\fi

\begin{abstract}

We propose a learning model for the task of visual storytelling. The main idea is to predict anchor word embeddings from the images and use the embeddings and the image features jointly to generate narrative sentences. We use the embeddings of randomly sampled nouns from the ground-truth stories as the target anchor word embeddings to learn the predictor.  To narrate a sequence of images, we use the predicted anchor word embeddings and the image features as the joint input to a seq2seq model. As opposed to state-of-the-art methods, the proposed model is simple in design, easy to optimize, and attains the best results in most automatic evaluation metrics. In human evaluation, the method also outperforms competing methods.

\end{abstract}

\vspace{-10pt}

\section{Introduction}

Visual storytelling, ie, narrating a sequence of images, is a challenging task~\cite{park2015expressing,huang2016visual}. It demands an understanding of the underlying storyline of the images. The process is naturally subjective. It often focuses more on conveying the narrator's own interpretation than describing the images in factual terms.

For example, as pointed out by the creators of the popular dataset VIST~\cite{huang2016visual}, concatenating the descriptions of the images does not give rise to a desirable narrative story. Table~\ref{tab:Diff_VIST_stats} illustrates the difference in the corpus statistics on the aforementioned dataset. Despite similar in length, story and caption use very different sets of words. At least 40\% of words that appear in stories do not appear in captions.

While this discrepancy has been well-documented, it is unclear how this insight could be used to devising effective models for visual storytelling. The task seems naturally gravitating to the method of \textsc{seq2seq} where we learn a mapping to encode a sequence of image features then to decode by outputting a sequence of words~\cite{gonzalez2018contextualize}. This method met some successes and is followed by others~\cite{huang2018hierarchically,wang2018show,wang2018no}.

\begin{table}[t]
    \small
    \centering
    \tabcolsep 15pt
    \begin{tabular}{l c c}
        \toprule
        & \bf Storys &  \bf Captions   \\
        \midrule
        Vocabulary Size    & 29,614 & 24,534 \\
        Avg. Sent. Length & 11.4 & 11.9 \\
        \midrule
        \# of Nouns         & 6,831 & 7,772 \\
        \# of Verbs        & 5,217 & 3,202 \\
        \# of Adjectives         & 2,089 & 2,018 \\
        \# of Adverbs        & 1,505 &    286 \\
        \bottomrule
    \end{tabular}
    \caption{Statistics of stories and captions on the VIST dataset~\cite{huang2016visual}}
    \vspace{-0.1in}
    \label{tab:Diff_VIST_stats}
    \vspace{-0.15in}
\end{table}

In this paper, we take a step toward identifying what might be needed for generating a narrative story. We hypothesize that each narrative story needs to have a sequence of \emph{anchor} words. For simplicity, we assume one anchor word per image. The anchor words form a prior on what can be ``said'' about the images. To narrate a sequence of images, our learning model just needs to predict the anchor word embedding for each image in turn and then supply the embeddings to a \textsc{seq2seq} model to generate the story.

But then, \emph{what are the anchor words}? They are not explicitly given in the annotated dataset. As a first step, we have shown that we can use the words in the ground-truth stories as anchor words and learn a predictive model (from the image features) to predict the anchor word embeddings when the ground-truth stories are not available.

As opposed to several best-performing models for the same task, our model is simple in design and does not need to use reinforcement learning to optimize~\cite{huang2018hierarchically,wang2018show,wang2018no}. Yet it attains the best performance in several evaluation metrics.

We describe the idea of using anchor words in section~\ref{sApproach}, supported by the evidence that such words, when added to a vanilla \textsc{seq2seq} model for story generation, significantly improve its performance. We then describe how to train a predictive model to predict its embedding.  In section~\ref{sExp}, we report our evaluation results and conclude in section~\ref{sConclusion}.

\vspace{-5pt}
\section{Related Work}

There is a large body of work in the intersection of vision and language, cf. ~\cite{lin2014microsoft,vinyals2015show}.

Image captioning is closely related to visual storytelling. \textsc{seq2seq} and its variants are among the most popular learning approaches for the task~\cite{xu2015show,vinyals2015show}.

From the very beginning, the creators of the dataset for visual storytelling highlighted the difference of captioning from narratives~\cite{huang2016visual}. In essence, narrative stories go beyond the factual enumeration of objects and activities depicted in the images, which is often adequate for image captioning.

Recent approaches for visual storytelling have been using reinforcement learning (RL) to optimize complicated models ~\cite{huang2018hierarchically,wang2018no}.  The approach proposed in this paper has the advantage of a simplified design and learning procedure, yet attains the best performance on several evaluation metrics.

\vspace{-5pt}
\begin{figure}[t]
    \centering
    \resizebox{\linewidth}{!}{
        \includegraphics[width=0.3\textwidth]{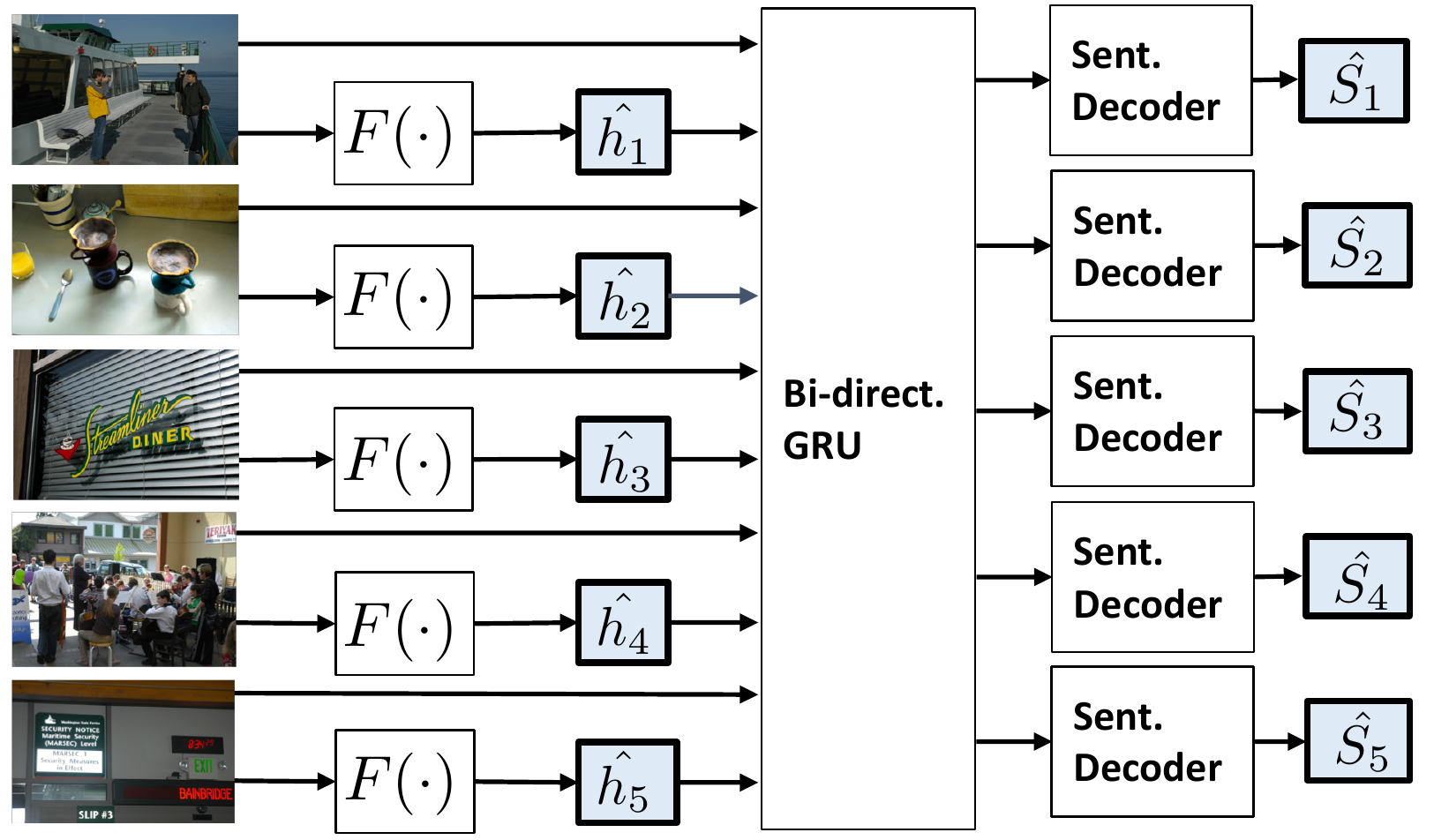}
    }
    \caption{Conceptual diagram of our approach for visual storytelling. The key difference from a typical \textsc{seq2seq} model is the component of predicting anchor word embeddings from the images. The predictions are then fused with the image features as the inputs for generating desired narrative sentences.}
    \label{fig:overall}
\end{figure}

\begin{table}[t]
    \small
    \centering
    \resizebox{\linewidth}{!}{
        \begin{tabular}{l c c c c}
            \toprule
            & B@4 & M & R & C  \\
            \midrule
            Image Only &  13.9 \tiny $\pm 0.4$ & 35.2 \tiny $\pm 0.1$ & 29.5 \tiny $\pm 0.3$ & 8.4 \tiny $\pm 0.9$\\
            \midrule
            \textbf{Anchoring}\\
            \  with {\bf Noun} & \textbf{17.2\tiny $\pm 0.2$} & \textbf{39.0\tiny $\pm 0.2$} & 33.8\tiny $\pm 0.1$ & \textbf{15.7\tiny $\pm 0.4$} \\
            \ with {\bf   Verb} & 16.5\tiny $\pm 0.1$ & 37.9\tiny $\pm 0.2$ & \textbf{34.7\tiny $\pm 0.2$} & 13.0\tiny $\pm 0.2$ \\
            \ with {\bf  Adj.} & 15.2\tiny $\pm 0.3$ & 36.6\tiny $\pm 0.1$ & 31.9\tiny $\pm 0.2$ & 11.3\tiny $\pm 0.2$ \\
            \ with {\bf   Adv.} & 14.9\tiny $\pm 0.3$ & 35.9\tiny $\pm 0.1$ & 31.0\tiny $\pm 0.2$ & 10.4\tiny $\pm 0.3$ \\
            \bottomrule
        \end{tabular}
    }
    \caption{Adding ground-truth words as anchor words to a \textsc{seq2seq} model significantly improves its performance where only image features are used. The higher numerical value indicates better performance.}
    \vspace{-0.2in}
    \label{tab:gt_anchor}
\end{table}

\section{Approach}
\label{sApproach}

The task of visual storytelling is to generate a sequence of narrative sentences $\{ \, \mathbf{S}_i, \, i=1, 2, \ldots, {N}\}$, one for each of the $N$ images $\{ \, \mathbf{I}_i, \, i=1, 2, \ldots, {N}\}$. The order of the images is important and is fixed. Each of the generated sentences $\mathbf{S}_i$ could contain a variable length of words.

The main idea behind our approach is straightforward. For each image, we learn and apply a model to predict its anchor word embedding. The predicted embedding is then concatenated with the image feature. The combined feature is fed into a \textsc{seq2seq}~\cite{sutskever2014sequence} where the narrative sentence is generated as output. Fig.~\ref{fig:overall} illustrates the model design.

The key challenge is to learn the anchor word prediction model when the dataset does not provide anchor words explicitly. We begin by describing how we overcome this challenge. Then we introduce our model in detail.

\subsection{What is an anchor word?}

We are inspired by the comparison between narrative stories and captions on the same sequence of images, shown in Table~\ref{tab:Diff_VIST_stats}.  In particular, a large number of words used in narration do not appear in captions. Intuitively speaking, they are less likely visually grounded.

Thus, we conjecture that possible candidates for anchor words are the words in the narrative sentences. The analysis in Table~\ref{tab:gt_anchor} confirms the usefulness of this hypothesis.

Specifically,  we train a model as in Fig.~\ref{fig:overall} with two variants. In the first variant, we supply only the image features. The results are reported in the row labeled as ``Image Only''.   In the second variant (``Anchoring''), we select all the noun (alternatively, verb, adjective, or adverb) words as anchor words -- one word per sentence in the story. We then train the \textsc{seq2seq} model by combining the image feature and the word embedding end to end.  The results are reported in rows labeled with the part-of-speech  (POS) tags of the selected words. For simplicity, all anchor words have the same POS tags. If there are multiple words with the same POS tags in each sentence, we randomly select one.

There are two points worth making. First, adding anchor words, irrespective of their types, significantly improves the performance of the \textsc{seq2seq} model with image features only. Note that the results in ``Image Only'' is on par with state-of-the-art results~\cite{wang2018no}. Secondly, among all POS tags, nouns as anchor words seem to be the most beneficial ones on all metrics except R(OUGE) where verbs improve more.

In the rest of this paper, we use nouns in the stories as the anchor words.

\begin{table*}[th]
    \small
    \centering
        \begin{tabular}{l c c c c c c c}
            \toprule
            Method & B@1 & B@2 & B@3 & B@4 & M & R & C  \\
            \midrule
            AREL \cite{wang2018no} & 63.8 & 39.1 & 23.2 & 14.1 & 35.0 & 29.5 & 9.4\\
            Show, Reward and Tell \cite{wang2018show} & 43.4 & 21.4 & 10.4 & 5.2 & - & - & 11.4 \\
            HSRL w/ Joint Training \cite{huang2018hierarchically} & - & - & - & 12.3 & 35.2& 30.8 & 10.7 \\
            \textsc{seq2seq}+Heuristics~\cite{huang2016visual} & - & - & - & - & 31.4 & - & -\\
            H-Attn-Rank\cite{yu2017hierarchically} & - & - & 20.8 & - & 33.9 & 29.8 & 7.4 \\
            \textbf{StoryAnchor}: Image Only & 62.2 \tiny $\pm2.5$ & 38.3 \tiny $\pm1.7$ & 22.7 \tiny $\pm0.8$ & 13.9 \tiny $\pm 0.4$ & 35.2 \tiny $\pm 0.1$ & 29.5 \tiny $\pm 0.3$ & 8.4 \tiny $\pm 0.9$ \\
            \textbf{StoryAnchor}: w/ Predicted Nouns & 65.1 \tiny $\pm 0.3$ & 40.0 \tiny $\pm 0.2$ & 23.4 \tiny $\pm 0$ & 14.0 \tiny $\pm 0.1$ & 35.5 \tiny $\pm 0.0$ & 30.0 \tiny $\pm 0.1$ & 9.9 \tiny $\pm 0.1$ \\
            \bottomrule
        \end{tabular}
    \caption{Comparison of  state-of-the-art method for the visual storytelling task on the VIST dataset. Our ``Image Only'' model is a reimplementation of XE+SS \cite{wang2018no} with the authors' public available codes.}
    \label{tab:SOTA}
    \vspace {-15pt}
\end{table*}

\subsection{Model and Learning}
\label{hwpred}

The data for our learning task is augmented with a list of anchor words $\{w_i \mid i=1, 2,\ldots, N\}$ corresponding to the images. Next, we explain how to learn each component.

\vspace{-10pt}

\paragraph{Anchor word embedding predictor} We learn a model $F(\mathbf{I}_i)$ to predict $w_i$. $F(\cdot)$ is parameterized by a  one-hidden-layer multi-layer perception (MLP) with ReLU non-linearity. The input could be the features for the $i$th image or all the images in the same sequence. In practice, there is no significant difference.

To be able to generalize to new anchor words, we predict its embedding and cast learning $F(\cdot)$ as a regression problem. To obtain the target (ie, the ``ground-truth'' embedding) for the word $w_i$, we take the embeddings from the ``Anchoring'' model in Table~\ref{tab:gt_anchor}. $F(\cdot)$ is then optimized to reduce the mean squared error between the predictions and the target anchor words embeddings.

\vspace{-10pt}

\paragraph{Story generation model} Similar to state-of-the-art visual storytelling methods~\cite{wang2018no,huang2016visual}, we use a \textsc{seq2seq} model~\cite{sutskever2014sequence} as story generator. Concretely, a bidirectional gated recurrent neural network~\cite{chung2014empirical}(GRU) is used to encode the concatenated feature of the image and the predicted anchor word embedding and to produce a sequence of hidden states
    \vspace{-5pt}
\begin{equation}
\vv_i = \text{BiGRU}(\mathbf{I}_i, F(\mathbf{I}_i))
    \vspace{-5pt}
\end{equation}
The sequence of the hidden states is then decoded by a one-layer GRU.  Both the encoder and the decoder are trained to maximize the likelihood of ground-truth stories.

\subsection{Other Implementation Details}

\paragraph{Visual and textual representation} We extracted the 2048 dimension feature from the penultimate layer of ResNet-152~\cite{he2016deep} as visual representations. The 512-dimensional word embedding is randomly initialized, which are fine-tuned in the training. Note that the anchor words are sharing the word embeddings with the words in the vocabulary.

\vspace{-10pt}

\paragraph{Model details} The concatenated features of the image and the anchor word embedding are projected into a 2048 dimensional feature with a one-hidden-layer MLP. Then, a one-layer BiGRU with 256-dimensional hidden states generates contextual embedding $\vv_i$ of 512 dimensions, to serve as hidden states representation. A standard \textsc{seq2seq} decoder with one-layer GRU with 512 hidden dimensions is used on top of these hidden states to generate a story.

\vspace{-10pt}

\paragraph{Optimization} As mentioned, the model is trained in two stages. In the first stage, ground-truth anchor words (nouns in the stories) are used to train the encoder-decoder as well as the embeddings end to end. The model is trained with mini-batches and ADAM for 100 epochs. Each mini-batch contains 64 sampled stories. The learning rate is initialized as 4e-4 and schedule sampling~\cite{bengio2015scheduled} has been used. The probability of schedule sampling is first set to be 0.05, increased by 0.05 every 5 epochs till 25 epochs. In the second stage, the predictor $F(\cdot)$ is trained. Specifically, we use the model that achieves the highest Meteor score on the validation set in the first stage training as a pre-trained model. We use the same optimization hyper-parameter to train the predictor with encoder-decoder model in an end-to-end way. The encoder-decoder and the word embeddings are kept fixed.

\paragraph{Inference} At the inference time (ie, narrating a sequence of images), we perform beam search for sentence decoding with a beam size of 3.

\section{Experiments}
\label{sExp}
\subsection{Experimental Setups}

\begin{table*}[bht]
    \small
    \centering
    \begin{tabular}{l c c c c c c c}
        \toprule
        & B@1 & B@2 & B@3 & B@4 & M & R & C  \\
        \midrule
        Human &  51.2 \tiny $\pm 0.2$  & 25.0 \tiny $\pm 0.2$ & 11.7 \tiny $\pm 0.2$  & 5.6 \tiny $\pm 0.2$ & 28.4 \tiny $\pm 0.1$ & 24.5 \tiny $\pm 0.1$  & 7.8 \tiny $\pm 0.1$  \\
        \midrule
        \textbf{StoryAnchor}: Image Only & 58.6 \tiny $\pm 0.2$  & 34.7 \tiny $\pm 0.2$ & 20.0 \tiny $\pm 0.1$  & 11.2 \tiny $\pm 0.1$ & 34.0 \tiny $\pm 0.1$ & 28.3 \tiny $\pm 0.1$  & 8.8 \tiny $\pm 0.1$ \\
        \textbf{StoryAnchor}: w/ Predicted Nouns & 60.7 \tiny $\pm 0.2$  & 35.8 \tiny $\pm 0.2$ & 20.3 \tiny $\pm 0.1$  & 11.9 \tiny $\pm 0.1$ & 34.5 \tiny $\pm 0.0$ & 28.9 \tiny $\pm 0.0$  & 10.1 \tiny $\pm 0.1$ \\
        \textbf{StoryAnchor}: w/ Ground-truth Nouns & 65.1 \tiny $\pm 0.2$  & 40.3 \tiny $\pm 0.1$ & 23.9 \tiny $\pm 0.1$  & 14.7 \tiny $\pm 0.0$ & 37.7 \tiny $\pm 0.1$ & 32.3 \tiny $\pm 0.1$  & 16.2 \tiny $\pm 0.1$ \\
        \bottomrule
    \end{tabular}
    \caption{Evaluating human performance by automatic evaluation procedures. Machine outperforms human in all metrics. }
    \label{tab:upperbound}
    \vspace{-15pt}
\end{table*}

 \begin{table}[t]
     \centering
          \small
         \begin{tabular}{l c c c}
             \toprule
                         & \textbf{StoryAnchor:} & AREL & Tie \\
                         &  w/ Predicted Nouns & & \\
             \midrule
                         Relevance & 53.2\%  & 40.4\% & 6.4\%\\
                         Concreteness & 45.1\%  & 38.1\% & 16.8\%\\
                         Coherence & 48.9\% &  42.3\% & 8.8\% \\
             \bottomrule
         \end{tabular}
     \caption{Human evaluation on the generated stories}
     \label{tab:abs_human}
         \vspace{-10pt}
 \end{table}

 \begin{table}[t]
      \centering
            \small
          \begin{tabular}{ c c c c}
              \toprule
                           \textbf{StoryAnchor:} & AREL & Human & Unsure \\
                              w/ Predicted Nouns & & \\
              \midrule
                     $19.9\%$ & $18.0\%$ & $\textbf{57.8\%}$ & $4.2\%$ \\
              \bottomrule
          \end{tabular}
      \caption{Which stories are preferred by human readers}
      \label{tab:abs_turing}
         \vspace{-10pt}
  \end{table}

\paragraph{Dataset} We use the VIST dataset~\cite{huang2016visual} for evaluation. It contains 10,032 visual albums with 50,136 stories. Each story contains five narrative sentences, corresponding to five grounded images respectively.

\paragraph{Evaluation} We follow the evaluation setup used in ~\cite{huang2016visual,wang2018no,yu2017hierarchically,huang2018hierarchically}. For each testing album, we sample one image sequence and generate a story based on that image sequence. The story is then scored against all 5 reference stories of that album. We use the evaluation code provided by the ~\cite{yu2017hierarchically}\footnote{\url{https://github.com/lichengunc/vist_eval}. This is the most commonly used evaluation script nowadays.}. We report results with average BLEU, METEOR, ROUGE, and CIDER over the test split.  We evaluate over 3 random runs and compute the means and variances of the metrics.

\paragraph{Identifying anchor words} We use NLTK POS tagger to get the tags. Each sentence contains on average 2.63 nouns, 2.0 verbs, 0.8 adjectives, and 0.5 adverbs. We use 'UNK' as the anchor word when there is no corresponding POS tag.

\subsection{Main Results}

We compare our method (\textit{StoryAnchor}) to several state-of-the-art methods~\cite{huang2016visual,yu2017hierarchically,wang2018no,wang2018show,huang2018hierarchically}. Figure~\ref{tab:SOTA} shows that our model performs significantly better than others in almost all evaluation metrics. In ROUGE and CIDER, approaches of using reinforcement learning seem to perform well.

We also conduct human evaluations to compare the outputs of our model and AREL~\cite{wang2018no}. We follow \cite{wang2018no} and design three questions to evaluate the relevance, concreteness, and coherence of generated stories and image sequences. 150 generated stories from the test splits are evaluated. For each story, 5 AMT workers are assigned. The reports are reported in Table \ref{tab:abs_human}. Our approach performs better.

\subsection{Analysis}

\paragraph{Is visual storytelling fundamentally out of reach of machines?} Are the metrics being used now to guide the design of our systems the right ones?

The results in Table~\ref{tab:upperbound} highlight the issues. There, we assess \emph{how well human storyteller would do}. For each album, we randomly select one human-written ground-truth story as ``generated'' story and the other 4 as ``reference'' stories. We then evaluate human performance by scoring the generated story. For a fair comparison, we re-evaluated all of the learning models with 4 sampled reference stories. Mean evaluation performances over five random runs are reported.

Clearly, the learning models outperform human storyteller significantly in every metric! Yet, our ``Turing test" suggests the opposite. In Table~\ref{tab:abs_turing}, over 450 stories (3 for each of the 150 sequences of images), we report the percentages of 150 AMT workers' preference of stories by two learning models and one human annotator. Human storytelling is much more preferred.  The misalignment between human evaluation and automatic evaluation metrics is likely a bottleneck for developing new methods for this task.

\vspace{-10pt}
\section{Conclusion}
\label{sConclusion}

The proposed StoryAnchor model is simpler in design. Yet, it attains the best results on most automatic evaluation metrics. The key insight is to use ``anchor words" to model the evolvement of the underlying storyline. Crudely, those words are the ``topics'' or ``states'' of the narrators. While those notions are not explicitly annotated in the current dataset, we have selected the nouns in the ground-truth stories as targets for learning an anchor word predictor.

{\small
\bibliographystyle{ieee}
\bibliography{egbib}
}

\end{document}